%% file: main.tex
\documentclass[runningheads]{llncs}
\usepackage{graphicx}
\usepackage[misc]{ifsym}
\usepackage{tikz}
\usepackage{comment}
\usepackage{amsmath,amssymb} 
\usepackage{color}

\usepackage[width=122mm,left=12mm,paperwidth=146mm,height=193mm,top=12mm,paperheight=217mm]{geometry}
\usepackage{hyperref}
\usepackage{comment}
\usepackage{subcaption}
\usepackage{multirow}

\usepackage{floatrow}
\newfloatcommand{figurebox}{figure}[\nocapbeside][\dimexpr(\textwidth-\columnsep)/2\relax]
\newfloatcommand{tablebox}{table}[\nocapbeside][\dimexpr(\textwidth-\columnsep)/2\relax]
\usepackage[utf8]{inputenc}
\usepackage{babel}
\usepackage[font=small,labelfont=bf]{caption}

\usepackage[accsupp]{axessibility}  

\begin{document}
\pagestyle{headings}
\mainmatter

\title{Continuous Spectral Reconstruction from RGB Images via Implicit Neural Representation} 


\titlerunning{Abbreviated paper title}
%








\author{Ruikang Xu\inst{1} \and
Mingde Yao\inst{1} \and Chang Chen\inst{2} \and
Lizhi Wang\inst{3} \and Zhiwei Xiong\inst{1}$^{(\textrm{\Letter})}$}
\authorrunning{R. Xu et al.}
\titlerunning{NeSR} 

%
\institute{University of Science and Technology of China \and
Huawei Noah's Ark Lab  \and Beijing Institute of Technology \\
\email{ \{xurk, mdyao\}@mail.ustc.edu.cn,}
\email{chenchang25@huawei.com,}
\email{wanglizhi@bit.edu.cn,}
\email{zwxiong@ustc.edu.cn}}
\maketitle

\begin{abstract}
Existing spectral reconstruction methods learn discrete mappings from spectrally downsampled measurements (e.g., RGB images) to a specific number of spectral bands.
However, they generally neglect the continuous nature of the spectral signature and only reconstruct specific spectral bands due to the intrinsic limitation of discrete mappings. 
In this paper, we propose a novel continuous spectral reconstruction network with implicit neural representation, which enables spectral reconstruction of arbitrary band numbers for the first time.
Specifically, our method takes an RGB image and a set of wavelengths as inputs to reconstruct the spectral image with arbitrary bands, where the RGB image provides the context of the scene and the wavelengths provide the target spectral coordinates. To exploit the spectral-spatial correlation in implicit neural representation, we devise a spectral profile interpolation module and a neural attention mapping module, which exploit and aggregate the spatial-spectral correlation of the spectral image in multiple dimensions.
Extensive experiments demonstrate that our method not only outperforms existing discrete spectral reconstruction methods but also enables spectral reconstruction of arbitrary and even extreme band numbers beyond the training samples.

\keywords{Computational photography, Hyperspectral image reconstruction, Implicit neural representation}
\end{abstract}

\input{1_Introduction}

\input{2_RelatedWork}
\input{3_Methods}

\input{4_Experiments}

\input{5_Conclusions}


\bibliographystyle{splncs04}
\bibliography{ref}
\end{document}

%% file: 1_Introduction.tex
\section{Introduction}
\label{sec:intro}

Spectral images record richer spectrum information than traditional RGB images, which have been proven useful in various vision-based applications, such as anomaly detection~\cite{jiang2021lren}, object tracking~\cite{xiong2020material}, and segmentation~\cite{dao2021improving}.
To acquire spectral images, conventional spectral imaging technology relies on either spatial or spectral scanning for capturing the spectral signature to a number of bands, which is of high complexity and time-consuming~\cite{descour1995computed,Wang_2015_CVPR,wang2016adaptive,cao2011prism,yao2019spectral}.
As an alternative, spectral reconstruction from RGB images is regarded as an attractive solution owning to the easy acquisition of RGB images~\cite{arad2016sparse,robles2015single,Sun_2021_CVPR,Zhu_2021_ICCV}.

\begin{figure}[!t]
  \centering
    \includegraphics[width=1\linewidth]{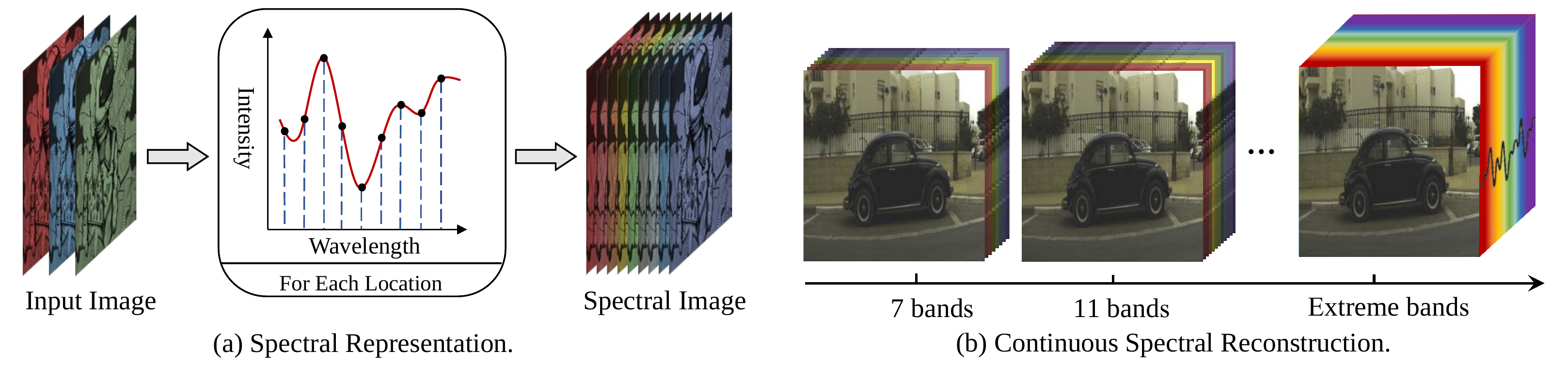}
   \caption{
  (a) The main difference of spectral reconstruction between previous methods and NeSR. Previous methods learn a discrete mapping to reconstruct the spectral image as the ``\textcolor{blue}{blue dashes}". We learn a continuous spectral representation and sample it to the target number of spectral bands as the ``\textcolor{red}{red curve}".
  (b) Continuoue spectral reconstruction.
   }
   \label{fig:teaser}
\end{figure}

Existing methods~\cite{shi2018hscnn+,li2020adaptive,zhao2020hierarchical,zhang2020pixel,sun2021tuning} for spectral reconstruction aim to learn a discrete mapping from an RGB image to a spectral image, which directly generates a specific number of spectral bands from three bands, as illustrated by the ``blue dashes" in Figure~\ref{fig:teaser} (a).
However, this kind of representation ignores the continuous nature of the spectral signature.
In the physical world, the spectral signature is in a continuous form where the high-dimension correlation is naturally hidden in the continuous representation~\cite{cao2010sparse,kuybeda2007rank}.
To approximate the natural representation of the spectral signature, we reformulate the spectral reconstruction process, where a number of spectral bands are resampled from a continuous spectral curve, as illustrated by the ``red curve" in Figure~\ref{fig:teaser} (a).

Recent works adopt the concept of implicit neural representation for super-resolution~\cite{chen2021learning} and 3D reconstruction~\cite{mildenhall2020nerf,sitzmann2019scene} tasks, which obtain the continuous representation of signals with high fidelity.
Inspired by this line of works, we aim to learn the continuous representation of the spectral signature for spectral reconstruction by leveraging a continuous and differentiable function~\cite{Niemeyer2020GIRAFFE,mescheder2019occupancy}.
However, it is non-trivial to exploit existing continuous representation methods for spectral images directly, since they only focus on the spatial correlation of RGB images~\cite{chen2021learning,sitzmann2020implicit}.
The high-dimension spatial-spectral correlation, which plays a vital role for high-fidelity spectral reconstruction, remains unexplored~\cite{Wang_2019_CVPR,wang2020super}.

To fill this gap, we propose Neural Spectral Reconstruction (NeSR) to continuously represent the spectral signature by exploiting the spatial-spectral correlation. 
Specifically, we first adopt a feature encoder to extract deep features from the input RGB image for representing the context information of the scene.
We then take the target wavelengths as the coordinate information to learn the projection from the deep features to the spectral intensities of the corresponding wavelengths.
To exploit the spatial-spectral correlation for continuous spectral reconstruction, we devise a Spectral Profile Interpolation (SPI) module and a Neural Attention Mapping (NAM) module. The former encodes the spatial-spectral correlation to the deep features leveraging the vertical and horizontal spectral profile interpolation, while the latter further enriches the spatial-spectral information using an elaborate spatial-spectral-wise attention mechanism.
With the above two modules, the spatial-spectral correlation is exploited in the deep features for learning a continuous spectral representation.

Benefiting from the continuous spectral representation as well as the exploitation of spatial-spectral correlation, the advantage of NeSR is twofold.
First, NeSR enables spectral reconstruction of arbitrary and even extreme band numbers beyond the training samples for the first time, as illustrated in Figure~\ref{fig:teaser} (b).
Second, NeSR notably improves the performance over state-of-the-art methods for spectral image reconstruction, bringing 12\% accuracy improvement with only 2\% parameter increase than the strongest baseline on the NTIRE2020 challenge dataset.

The contributions of this paper are summarized as follows:

\textbf{(1)}
For the first time, we propose NeSR to reconstruct spectral images with an arbitrary number of spectral bands while keeping high accuracy via implicit neural representation.

\textbf{(2)}
We propose the SPI and NAM modules to exploit the spatial-spectral correlation of depth features extracted from input RGB images for continuous spectral representation.

\textbf{(3)}
Extensive experiments demonstrate that NeSR outperforms state-of-the-art methods in reconstructing spectral images with a specific number of bands.

%% file: 2_RelatedWork.tex
\section{Related Work}
\label{sec:Related}

\subsection{Implicit Neural Representation}
Implicit neural representation aims to model an object as a continuous and differentiable function that maps coordinates and deep features to the corresponding signal, which is parameterized by a deep neural network.
Recent works demonstrate its potential for modeling surfaces~\cite{chabra2020deep,jiang2020local,sitzmann2019scene}, shapes~\cite{atzmon2020sal} \cite{michalkiewicz2019implicit}, and the appearance of 3D objects~\cite{niemeyer2020differentiable,oechsle2019texture}.
Mildenhall~\emph{et al.} first introduce the implicit neural representation for synthesizing novel views of complex scenes using a sparse set of input views, named Neural Radiance Filed (NeRF)~\cite{mildenhall2020nerf}.
Compared with explicit 3D representations, implicit neural representation can capture subtle details.
With the great success of NeRF~\cite{mildenhall2020nerf}, a large number of works extend implicit neural representation to other applications, such as 3D-aware generalization~\cite{tancik2021learned,Chan_2021_CVPR}, pose estimation~\cite{yen2020inerf,su2021nerf}, relighting~\cite{srinivasan2021nerv,boss2021nerd}.

To get a more general representation, recent works estimate latent codes and share an implicit space for different objects or scenes~\cite{chen2021learning,2021Implicit,sitzmann2020implicit,chen2019learning}, instead of learning an independent implicit neural representation for each object.
The implicit space can be generated with an auto-encoder architecture.
For example, Sitzmann~\emph{et al.}~\cite{sitzmann2020metasdf} propose a meta-learning-based method for sharing the implicit space.
Mescheder~\emph{et al.}~\cite{mescheder2019occupancy} propose to generate a global latent space of given images as input and use an occupancy function conditioning to perform the 3D reconstruction.
Sitzmann~\emph{et al.}~\cite{sitzmann2020implicit} replace ReLU with periodic activation functions and demonstrate that it can model the signals in higher quality.
Chen~\emph{et al.}~\cite{chen2021learning} utilize local implicit image function for representing natural and complex images.
Yang~\emph{et al.}~\cite{2021Implicit} propose a Transformer-based implicit neural representation for screen image super-resolution.
Different from these works, we propose NeSR to address the continuous spectral reconstruction from RGB images by fully exploring the spatial-spectral correlation.

\subsection{Spectral Reconstruction from RGB Images}
Most existing spectral imaging systems rely on either spatial or spectral scanning~\cite{descour1995computed,Wang_2015_CVPR,wang2016adaptive}.
They need to capture the spectral information of a single point or a single band separately, and then scan the whole scene to get a full spectral image.
Due to the difficulty of measuring information from scenes with moving content, they are unsuitable for real-time operation. Moreover, these spectral imaging systems remain prohibitively expensive for consumer-grade usage.
As an alternative, reconstructing spectral images from RGB images is a relatively low-cost and convenient approach to acquire spectral images.
However, it is a severely ill-posed problem, since much information is lost after integrating the spectral radiance into RGB values.

Many methods have been proposed for spectral reconstruction from RGB images~\cite{shi2018deep,wang2018high}.
Early works leverage sparse coding to recover the lost spectral information from RGB images.
Arad~\emph{et al.}~\cite{arad2016sparse} first exploit the spectral prior from a vast amount of data to create a sparse dictionary, which facilitates the spectral reconstruction.
Later, Aeschbacher~\emph{et al.}~\cite{aeschbacher2017defense} further improve the performance of Arad’s method leveraging the A+ framework~\cite{timofte2014a+} from super-resolution.
Alternatively, Akhtar~\emph{et al.}~\cite{akhtar2018hyperspectral} utilize Gaussian processes and clustering instead of dictionary learning.
With the great success of convolution neural networks in low-level computer vision, this task has received increasing attention from the deep learning direction~\cite{Zhu_2021_ICCV,zhang2020pixel,Sun_2021_CVPR}.
Xiong~\emph{et al.}~\cite{xiong2017hscnn} propose a unified deep learning framework HSCNN for spectral reconstruction from both RGB and compressive measurements.
Shi~\emph{et al.}~\cite{shi2018hscnn+} propose two improvements to boost the performance of HSCNN.
Li~\emph{et al.}~\cite{li2020adaptive} propose an adaptive weighted attention network (AWAN) to explore the camera spectral sensitivity prior for improving the reconstruction accuracy. 
Zhao~\emph{et al.}~\cite{zhao2020hierarchical} propose a hierarchical regression network (HRNet) for reconstructing spectral images from RGB images.
Cai~\emph{et al.}~\cite{cai2022mask} propose a Transformer-based model for spectral reconstruction, which utilize the mask-guided attention mechanism to capture the global correlation for reconstruction.
However, these methods reconstruct the spectral images by the discrete mapping, which fixes the number of output spectral bands and ignores the continuous nature of the spectral signature.
In this work, we propose NeSR to learn the continuous representation for spectral reconstruction from RGB images, which improves accuracy for reconstructing spectral images with a specific number of bands and enables spectral reconstruction with an arbitrary and extreme number of bands.

%% file: 3_Methods.tex
\section{Neural Spectral Reconstruction}

\begin{figure*}[!t]
  \centering
  \includegraphics[width=1\linewidth]{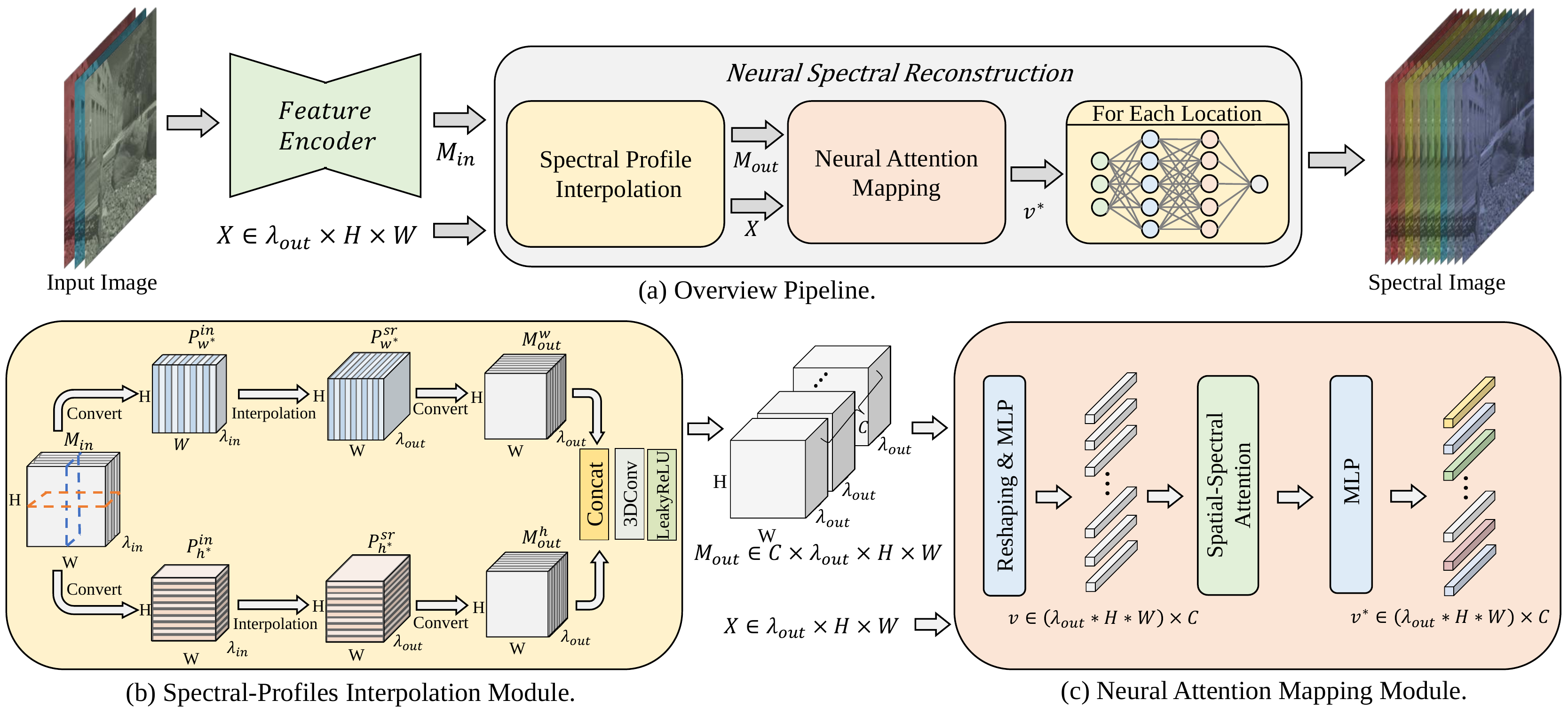}
  \caption{The proposed Neural Spectral Reconstruction.
  }
  \label{fig:nesr}
\end{figure*}

\subsection{Overview}

The overview of NeSR is shown in Figure~\ref{fig:nesr} (a).
Given an input RGB image, we cascade a feature encoder to extract the deep feature $M_{in} \in \mathbb{R}^{\lambda_{in} \times H \times W}$ as one input.
For learning the continuous spectral representation, we take the 3D coordinate $X \in \mathbb{R}^{\lambda_{out} \times H \times W}$ as the other input.
Specifically, for the spectral dimension, we use the normalized spectral wavelengths to reconstruct the spectral intensities of the corresponding wavelengths. 
Then, NeSR takes the deep feature $M_{in}$ and the coordinate $X$ to reconstruct the spectral image with the target number of bands as the output, denoted as
\begin{equation}
\begin{aligned}
Y = \mathcal{F}(M_{in}, X),
\end{aligned}
\end{equation}
where $\mathcal{F}$ stands for the overall processing of NeSR.

Specifically, given the deep feature $M_{in}$ and the coordinate $X$, NeSR upsamples the deep feature to the target number of spectral bands for subsequent coordinate information fusion at first.
To this end, the SPI module is designed to upsample the deep feature and encode the spatial-spectral correlation leveraging the vertical and horizontal spectral profile interpolation.
Then, the NAM module is designed to further exploit the spatial-spectral correlation of the fused feature by global attention in spatial and spectral dimensions, which computes the similarity of different channels according to the spatial-spectral-wise attention mechanism.

After exploiting the spatial-spectral correlation and fusing the coordinate information, a new deep feature $v^{*}$ is generated.
Finally, each latent code of this feature is fed to a multi-layer perception (MLP) to predict the corresponding spectral signature. Iterating over all locations of the target spectral image, NeSR can reconstruct the spectral image with the desired number of bands while keeping high reconstruction fidelity.

\subsection{Spectral Profile Interpolation}

The SPI module upsamples the deep feature to the target number of spectral bands and encodes the spatial-spectral correlation of the spectral image. The spatial-spectral correlation is an essential characteristic of spectral representation~\cite{Wang_2019_CVPR,wang2020super}.
To upsample the feature while exploiting this correlation, we propose the concept of Spectral Profile (SP) inspired by other high-dimension image reconstruction tasks~\cite{zuckerman2020across,8022901,xiao2020space}. 

We first give the definition of SP as follows.
Consider a spectral image as a 3D volume $I(h, w, \lambda)$, where h and w stands for the spatial dimensions and $\lambda$ stands for the spectral dimension, the vertical SP $P_{w^*}(h,\lambda)$ and the horizontal SP $P_{h^*}(w,\lambda)$ are the slices generated when $w = w^*$ and $h = h^*$, respectively.
Since much information is lost after integrating the spectral radiance into RGB values, we exploit the spatial-spectral correlation of the spectral image in the feature domain.
The flow diagram of the SPI module is shown in Figure~\ref{fig:nesr} (b).

In the SPI module, given the deep feature $M_{in} \in \mathbb{R}^{\lambda_{in} \times H \times W}$, we first convert it into the vertical and horizontal SPs to exploit the correlations in the spatial and spectral dimensions, denoted as $P^{in}_{w^*}(h,\lambda) \in \mathbb{R}^{H \times \lambda_{in}}$ and $P^{in}_{h^*}(w,\lambda) \in \mathbb{R}^{W \times \lambda_{in}}$.
Then, the upsampled vertical SP $P^{sr}_{w^*}(h, \lambda) \in \mathbb{R}^{H \times \lambda_{out}}$ and the upsampled horizontal SP $P^{sr}_{h^*}(w, \lambda) \in \mathbb{R}^{W \times \lambda_{out}}$ can be generated as
\begin{equation}
\begin{aligned}
P^{sr}_{w^*}(h, \lambda) = Up \left(P^{in}_{w^*}(h,\lambda)\right), \\
P^{sr}_{h^*}(w, \lambda) = Up \left(P^{in}_{h^*}(w,\lambda)\right),
\end{aligned}
\end{equation}
where $Up(\cdot)$ stands for the upsampling operation.
After converting the above upsampled SPs back to the feature maps, we obtain the intermediate results $M^{h}_{out}$, $M^{w}_{out}$ $\in \mathbb{R}^{\lambda_{out} \times H \times W}$.
To generate the upsampled deep feature for the coordinate information fusion, we fuse the intermediate results with the concatenation operation and utilize the 3D convolution layers with LeakyReLU, which can be denoted as
\begin{equation}
\begin{aligned}
M_{out} = Conv_{3D}\left(\left[M^{h}_{out}, M^{w}_{out}\right]\right),
\end{aligned}
\end{equation}
where $M_{out} \in \mathbb{R}^{C \times \lambda_{out} \times H \times W}$ stands for the output feature of the SPI module, which can be viewed as $\lambda_{out} \times H \times W$ latent codes corresponding to the coordinate of the spectral image.
$Conv_{3D}(\cdot)$ and $[\cdot, \cdot]$ denote the 3D convolution layers with LeakyReLU and the concatenation operation, respectively.

\subsection{Neural Attention Mapping}

The NAM module further exploits the spatial-spectral correlation for continuous spectral representation by a new attention mechanism.
Inspired by the success of self-attention-based architectures, such as Transformer, in natural language processing~\cite{vaswani2017attention} and computer vision~\cite{weng2021event,chen2021pre,zhu2020deformable}, we propose the spatial-spectral-wise attention mechanism, which captures the interactions of different dimensions to exploit the spectral-spatial correlation.
The flow diagram of the NAM module is shown in Figure~\ref{fig:nesr} (c).

Different from the standard Transformer block that receives a 1D token embedding~\cite{dosovitskiy2020image} as input, the NAM module takes a 3D tensor as input.
To handle the 3D tensor, we first reshape the 3D feature $M_{out}$ and the 3D coordinate $X$ to the 1D token embedding, and then map the embedding with an MLP, denoted as
\begin{equation}
\begin{aligned}
v= MLP \left(Reshape([M_{out}, X])\right),
\end{aligned}
\end{equation}
where $MLP(\cdot)$ and $Reshape(\cdot)$ stand for the MLP and the reshape operation, respectively.
$v \in \mathbb{R}^{(\lambda_{out} \cdot H \cdot W) \times C}$ denotes the fused token embedding, which fuses the information of the coordinate and the deep feature.
Then, $v$ is fed to the spatial-spectral attention block to exploit the spatial-spectral correlation of the fused token embedding.

In computing attention, we first map $v$ into key and memory embeddings for modeling the deep correspondences.
Unlike the self-attention mechanism that only focuses on a single dimension, we compute the attention maps of different dimensions to capture the interactions among dimensions for exploring the correlation of the fused token embedding.
Benefiting from that $v$ compresses the spectral-spatial information to one dimension, the attention maps fully exploit the spectral-spectral correlation to compute the similarity among channels, denoted as
\begin{equation}
\begin{aligned}
Q, K, V &= vW_{q},  vW_{k}, vW_{v},\\
v^{*} &= MLP(V \otimes {\rm SoftMax}(Q^{\rm T} \otimes K)),
\end{aligned}
\end{equation}
where $Q, K, V \in \mathbb{R}^{(\lambda_{out} \cdot H \cdot W) \times C}$ stand for \emph{query}, \emph{key} and \emph{value} to generate the attention, and $W_{q},W_{k}, W_{v}$ stand for the corresponding weights, respectively.
$\otimes$ and $T$ denote the batch-wise matrix multiplication and the batch-wise transposition, respectively.
Finally, we generate the output token embedding of the NAM module $v^{*} \in \mathbb{R}^{(\lambda_{out} \cdot H \cdot W) \times C}$ by an MLP.

\subsection{Loss Function}
We adopt the Mean Relative Absolute Error (MRAE) as the loss function, which is defined as
\begin{equation}
\mathcal{L} = \frac{1}{N} \sum_{i=1}^{N}(Y^{(i)}-I_{GT}^{(i)} /(I_{G T}^{(i)}+\varepsilon)),
\end{equation}
where $Y^{(i)}$ and $I_{GT}^{(i)}$ stand for the $i^{th}$ $(i = 1, . . . , N)$ pixel of the reconstructed and ground truth spectral images, respectively.
We set $\varepsilon=1 \times 10^{-3}$ due to zero points in the ground truth spectral image.

%% file: 4_Experiments.tex
\section{Experiments}\label{sec:exp}

\begin{table*}[!t]

  \renewcommand{\arraystretch}{1.15} 

    \resizebox{0.99\linewidth}{!}{
\begin{tabular}{cccccccccccc}

\hline\hline
  \multirow{2}{*}{Methods} &
  \multicolumn{2}{c}{NTIRE2020 Clean} &
  \multicolumn{2}{c}{NTIRE2020 Real} &
  \multicolumn{2}{c}{CAVE} &
  \multicolumn{2}{c}{NTIRE2018 Clean} &
  \multicolumn{2}{c}{NTIRE2018 Real} &
  \multirow{2}{*}{Param (M)} \\ \cline{2-11} 
     & MRAE    & RMSE     & MRAE    & RMSE      & MRAE   &    RMSE    & MRAE       & RMSE     & MRAE     & RMSE     \\ \hline
BI  & 0.16566  & 0.04551 & 0.17451  & 0.04307   & 5.74425  & 0.16886 & 0.12524  & 0.01941  & 0.15862  & 0.02375 & -  \\ \hline
HSCNN-R          & 0.04128 & 0.01515 & 0.07278 & 0.01892 & 0.19613 & 0.03535 & 0.01943 & 0.00389 & 0.03437 & 0.00621 & 1.20 \\
HSCNN-R + NeSR    & 0.03832 & 0.01323 & 0.06978  & 0.01814  & \underline{\color{blue}{0.15353}} & 0.03304 
& 0.01611 & 0.00383 & 0.03171 & 0.00584 & 1.34 \\ \hline
HRNet          & 0.04007 & 0.01401 & 0.06756 & 0.01821  & 0.17219 & 0.02987
& 0.01521 & 0.00368
& \underline{\color{blue}{0.02985}} &\underline{\color{blue}{0.00571}} & 31.70\\
HRNet + NeSR   & 0.03701 & 0.01284 &\underline{\color{blue}{0.06691}} & 0.01797 
& 0.16431 & \underline{\color{blue}{0.02977}}
& 0.01466 & 0.00351 
& \textbf{\color{red}{0.02895}} 
& \textbf{\color{red}{0.00557}}
 & 31.88\\ \hline
AWAN           & \underline{\color{blue}{0.03441}} & \underline{\color{blue}{0.01215}} 
& 0.06883 & \underline{\color{blue}{0.01711}} 
& 0.19156  & 0.03752
& \underline{\color{blue}{0.01226}} 
& \underline{\color{blue}{0.00255}}
& 0.03121 & 0.00577 & 28.59 \\
AWAN + NeSR   & \textbf{\color{red}{0.02996}} 
& \textbf{\color{red}{0.00989}} 
&\textbf{\color{red}{0.06661}} 
& \textbf{\color{red}{0.01632}}
& \textbf{\color{red}{0.13226}}
& \textbf{\color{red}{0.02789}}
& \textbf{\color{red}{0.01159}} 
& \textbf{\color{red}{0.00229}} 
& 0.03003 & 0.00552 & 29.29 \\ \hline\hline
\end{tabular}
}
\caption{Quantitative comparison on the NTIRE2020, CAVE and NTIRE2018 datasets for 31-band spectral reconstruction from RGB images.
  \textbf{\color{red}{Red}} and \underline{\color{blue}{blue}} indicate the best and the second best performance, respectively. }
  \label{tab:4_exp_main1}
\end{table*}%


\subsection{Comparison to State-of-the-art Methods}\label{sota}
To quantitatively evaluate the effectiveness of the proposed method, we first compare NeSR with state-of-the-art methods in reconstructing spectral images with a specific number of spectral bands.

\begin{figure}[t]

\includegraphics[width=1\linewidth]{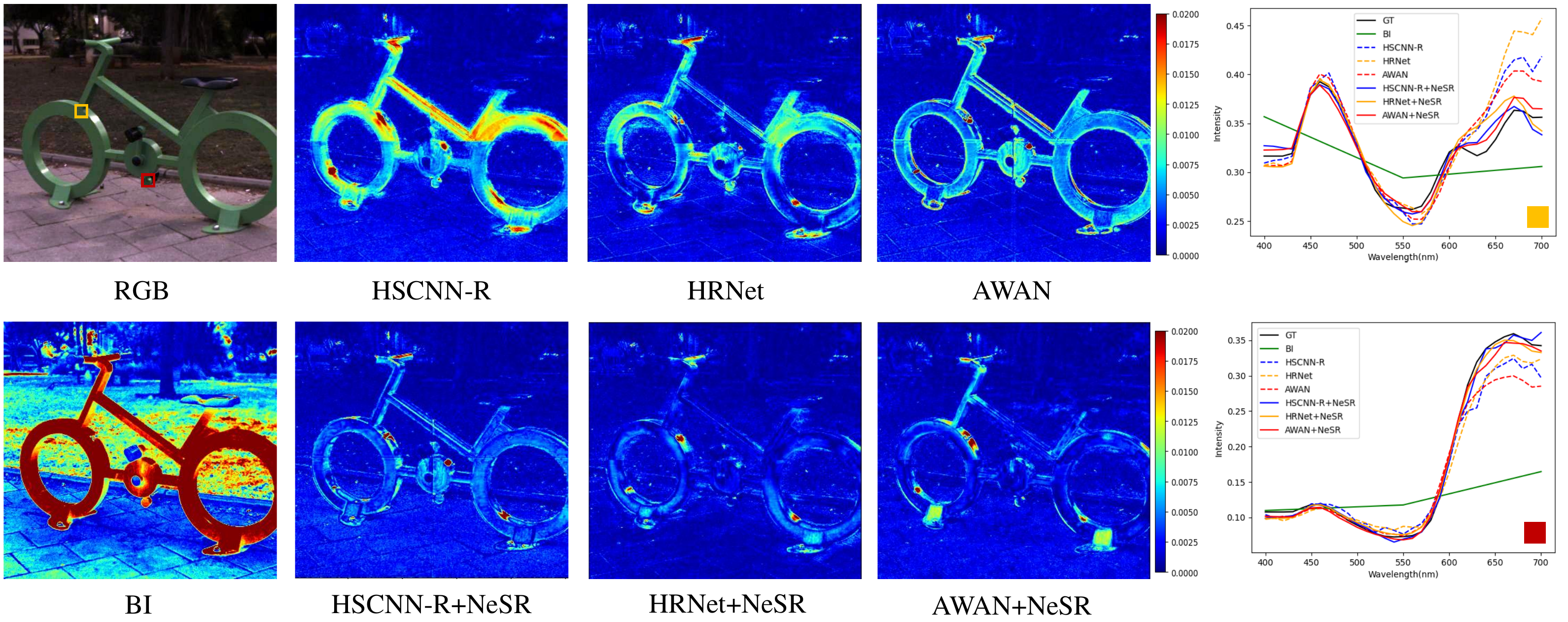}

\caption{Visualization of the error maps of different methods of spectral reconstruction from RGB images on ``ARAD\_HS\_0465'' of the NTIRE2020 ``Clean"  dataset. We show the spectral signatures of selected pixels and mark them by red and yellow rectangles in the RGB image. Please zoom in to see the difference of spectral intensity.}  
\label{fig:compare12}
\end{figure}

\textbf{Datasets.} In this work, we select three datasets as the benchmark for training and evaluation, including NTIRE2020~\cite{arad2020ntire}, CAVE~\cite{yasuma2010generalized} and NTIRE2018~\cite{arad2018ntire}.
NTIRE2020 and NTIRE2018 are the benchmarks for the hyperspectral reconstruction challenges in NTIRE2020 and NTIRE2018, respectively.
Both datasets consist of two tracks: ``Clean" and ``Real World", in which each spectral image consists of 31 successive spectral bands ranging from 400 nm to 700 nm with a 10 nm increment.
CAVE is a 16-bit spectral image dataset containing 32 scenes with 31 successive spectral bands ranging from 400 nm to 700 nm with a 10 nm increment.
For training, the image pairs of the three datasets are randomly cropped to $64 \times 64$ and normalized to range $[0, 1]$.
The training sets of the NTIRE2020 and NTIRE2018 datasets and the randomly picked 28 scenes from the CAVE dataset are used for training.
We use the validation sets of the NTIRE2020 and NTIRE2018 datasets and the remaining 4 scenes from the CAVE dataset as the test datasets.
We utilize the MRAE and Root Mean Square Error (RMSE) as metrics to quantitatively evaluate the reconstructed spectral images.

\textbf{Implementation Details.}
Our proposed method is trained on spectral-RGB images on the three datasets separately.
We take $\lambda_{min}=400$ nm and $\lambda_{max}=700$ nm as the minimum and maximum spectral coordinates since the input RGB images are synthesized in such a visible light range.
We then normalize the target spectral coordinate $\lambda_i$ to [-1, 1] as $2 \times (\lambda_i - \lambda_{min})/(\lambda_{max}-\lambda_{min})-1$.
We employ a 4-layer MLP with ReLU activation to reconstruct spectral images at the end of NeSR, and the hidden dimensions are (128, 128, 256, and 256).
We take the bilinear interpolation as the upsampling operation in the SPI module. 
Adam optimizer is utilized with parameters $\beta_1 = 0.9$ and $\beta_2 = 0.999$.
The learning rate is initially set to $1 \times 10^{-4}$ and is later downscaled by a factor of 0.5 after every $2 \times 10^{4}$ iterations till $3 \times 10^{5}$ iterations.
All the experiments in this paper are conducted in PyTorch 1.6.

\textbf{Comparison Methods.}
To verify the superiority of our method, we compare NeSR with different baselines, including one classical method (bilinear interpolation, BI) and three deep-learning-based methods (HSCNN-R~\cite{shi2018hscnn+}, HRNet~\cite{zhao2020hierarchical} and AWAN~\cite{li2020adaptive}).
We retrain the baselines under the same configuration of our method for a fair comparison.
In NeSR, we set the feature encoder as one of the aforementioned deep-learning-based methods by removing the output layer, respectively.

\textbf{Quantitative Evaluation.} 
Table~\ref{tab:4_exp_main1} shows the results of different methods on all benchmarks.
It can be seen that, on the one hand, NeSR boosts the performance of the corresponding baseline with a slight parameter increase.
On the other hand, NeSR is architecture-agnostic to be plugged into various backbones,  which is effective for both simple (e.g., HSCNN-R) and complex (e.g., AWAN) architectures.
Specifically, HRNet+NeSR achieves 6.9\% decrease in MRAE than HRNet with only 0.5\% parameters increase on the validation set of NTIRE2020 ``Clean", and it shows the best performance on the validation set of NTIRE2018 ``Real World".
AWAN+NeSR achieves the best performance on the validation set of two tracks of NTIRE2020, which brings 12.9\% decrease in MRAE than AWAN with 2.4\% parameters increase.

\textbf{Qualitative Evaluation.} 
To evaluate the perceptual quality of spectral reconstruction, we show the error maps of different methods on the NTIRE2020 ``Clean" validation set.
As visualized in Figure~\ref{fig:compare12},  we can observe a significant error decrease brought by NeSR.
Besides, the spectral signatures reconstructed by NeSR for selected pixels are closer to the ground truth than the corresponding baselines (the yellow and red boxes).
Therefore, NeSR can boost all the corresponding baselines in terms of spatial and spectral reconstruction accuracy.

\begin{table}[!t]
 \small

      \renewcommand{\arraystretch}{1.1} 

      \resizebox{0.99\linewidth}{!}{
\begin{tabular}{ccccccccc}
\hline\hline
\multirow{2}{*}{Methods} &

  \multicolumn{2}{c}{31 bands } &
  \multicolumn{2}{c}{16 bands } &
  \multicolumn{2}{c}{11 bands } &
  \multicolumn{2}{c}{7 bands } \\  \cline{2-9}
                                  & MRAE    & RMSE     & MRAE    & RMSE     & MRAE    & RMSE     & MRAE    & RMSE     \\ \hline 
BI   & 0.15573  & 0.02941  & 0.16611  & 0.03052  & 0.19012  & 0.03293  & 0.20214  & 0.03276  \\ 
Sparse coding & 0.04843 & 	0.01116	& 0.05722 &	0.01287 & 0.07681 &	0.01451 & 0.09841 & 0.02079
 \\ \hline 
AWAN (-D)  & 0.02298 & 0.00534 & 0.02313 & 0.00544 & 0.02412 & 0.00565 & 0.02535 & 0.00578 \\

AWAN (-S) & \underline{\color{blue}{0.02212}} & \underline{\color{blue}{0.00530}} & \underline{\color{blue}{0.02241}} & \underline{\color{blue}{0.00543}} & \underline{\color{blue}{0.02347}} & \underline{\color{blue}{0.00567}} & \underline{\color{blue}{0.02484}} & \underline{\color{blue}{0.00571}}
\\ \hline

AWAN + NeSR & \textbf{\color{red}{0.02001}} & \textbf{\color{red}{0.00498}} & \textbf{\color{red}{0.02024}} & \textbf{\color{red}{0.00505}} & \textbf{\color{red}{0.02129}} & \textbf{\color{red}{0.00522}} & \textbf{\color{red}{0.02119}} & \textbf{\color{red}{0.00541}}

\\ 
\hline\hline
\end{tabular}

}
\caption{Quantitative comparison on the ICVL dataset for spectral reconstruction from RGB images with arbitrary bands.
\textbf{\color{red}{Red}} and \underline{\color{blue}{blue}} indicate the best and the second best performance, respectively.}
\label{table2}
\end{table}%

\begin{figure}[!t]
  \centering
  \includegraphics[width=1\linewidth]{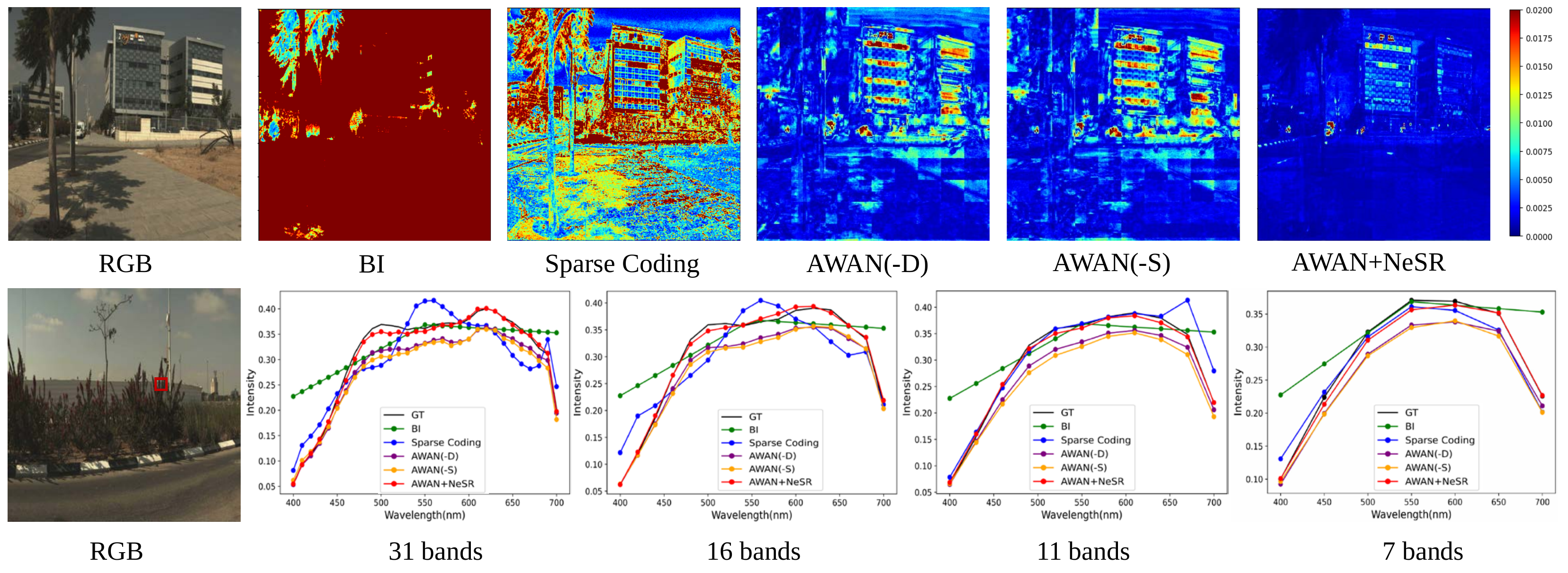}
  \caption{Top: Visualization of the error maps of different methods of spectral reconstruction from RGB images on ``BGU\_HS\_00044" of the ICVL dataset with 31 bands.
  Bottom: Spectral curves of reconstructed images of spectral reconstruction from RGB images with different band numbers on ``BGU\_HS\_00045" of ICVL.
  }
  \label{fig:compare1}
\end{figure}

\subsection{Continuous Spectral Reconstruction}

Previous methods of spectral reconstruction fix the number of output bands.
Different from existing methods, by leveraging the continuous spectral representation, NeSR can reconstruct the spectral images with an arbitrary number of spectral bands and maintain high fidelity using a single model.

\textbf{Dataset and Implementation.} We build a spectral dataset based on ICVL dataset \cite{arad2016sparse} for \textit{continuous spectral reconstruction}.
We remove duplicate scenes from the ICVL dataset and randomly choose 80 scenes for training and 10 scenes for testing.
We process raw data of the ICVL dataset to generate the spectral images from $400$ nm to $700$ nm with different band numbers.
Specifically, 31/16/11/7 spectral bands are used to validate the effectiveness of NeSR, which can be replaced by other band numbers.
The implementation details are the same as in Section~\ref{sota}.

\textbf{Comparison Methods.} Since there are few available methods that could reconstruct spectral images with different bands using a single model, we set up three methods as baselines in this section (See Table~\ref{table2}).
BI: We interpolate RGB images following the prior work~\cite{zhang2020pixel,Zhu_2021_ICCV}, which operates the interpolation function along the spectral dimension, to obtain spectral images with the desired band numbers.
Sparse coding: The sparse coding method~\cite{arad2016sparse} reconstructs spectral images leveraging a sparse dictionary of spectral signatures and their corresponding RGB projections, which can reconstruct spectral images with desired output band numbers.
Deep-learning-based: We select AWAN~\cite{li2020adaptive} as a representative deep-learning-based baseline for comprehensive comparisons.
Because the current deep models cannot reconstruct spectral images with an arbitrary number of bands, we separately train different models for different band numbers (7/11/16/31 bands), denoted as AWAN(-S).
We also design a two-step strategy, denoted as AWAN(-D), by first reconstructing spectral images with a large number of spectral bands (e.g., 61 bands) and then downsampling them to the target number of spectral bands.

\textbf{Quantitative Evaluation.}
We compare the quantitative results between NeSR and the baselines as mentioned above.
As shown in Table~\ref{table2}, reconstructing the spectral image with an arbitrary number of bands by BI shows poor performance because the RGB image is integrated on the spectral dimension, and direct interpolation between RGB channels cannot correspond to any spectral wavelength.
The sparse coding method \cite{arad2016sparse} also has a limited performance since it relies on the sparse dictionary but lacks the powerful representation of the deep neural networks.
Our continuous spectral representation (AWAN+NeSR) can reconstruct spectral images into arbitrary bands with a single model, which surpasses the aforementioned methods.
Compared with AWAN(-S), NeSR achieves 9.5\%, 9.6\%, 9.2\%, and 14.5\% decrease in terms of MRAE for reconstructing 31/16/11/7 bands.
The comparison results demonstrate that NeSR can effectively reconstruct spectral images with an arbitrary number of bands leveraging the continuous spectral representation.

\textbf{Qualitative Evaluation.}
To give a visual comparison of different methods, we display the error maps of one representative scene from the ICVL dataset with 31 spectral bands in Figure~\ref{fig:compare1}.
From the visualization, we can see that NeSR provides higher reconstruction fidelity than other methods.
We also give the spectral curves of different methods for reconstructing different spectral band numbers, which shows that our method has a lower spectral error than other methods.
The qualitative comparison demonstrates that NeSR can reconstruct spectral images with an arbitrary number of spectral bands while keeping high fidelity.

\subsection{Extreme Spectral Reconstruction}\label{subsec:out_of_distribution}

\begin{figure}[!t]

  \centering
  \includegraphics[width=0.85\linewidth]{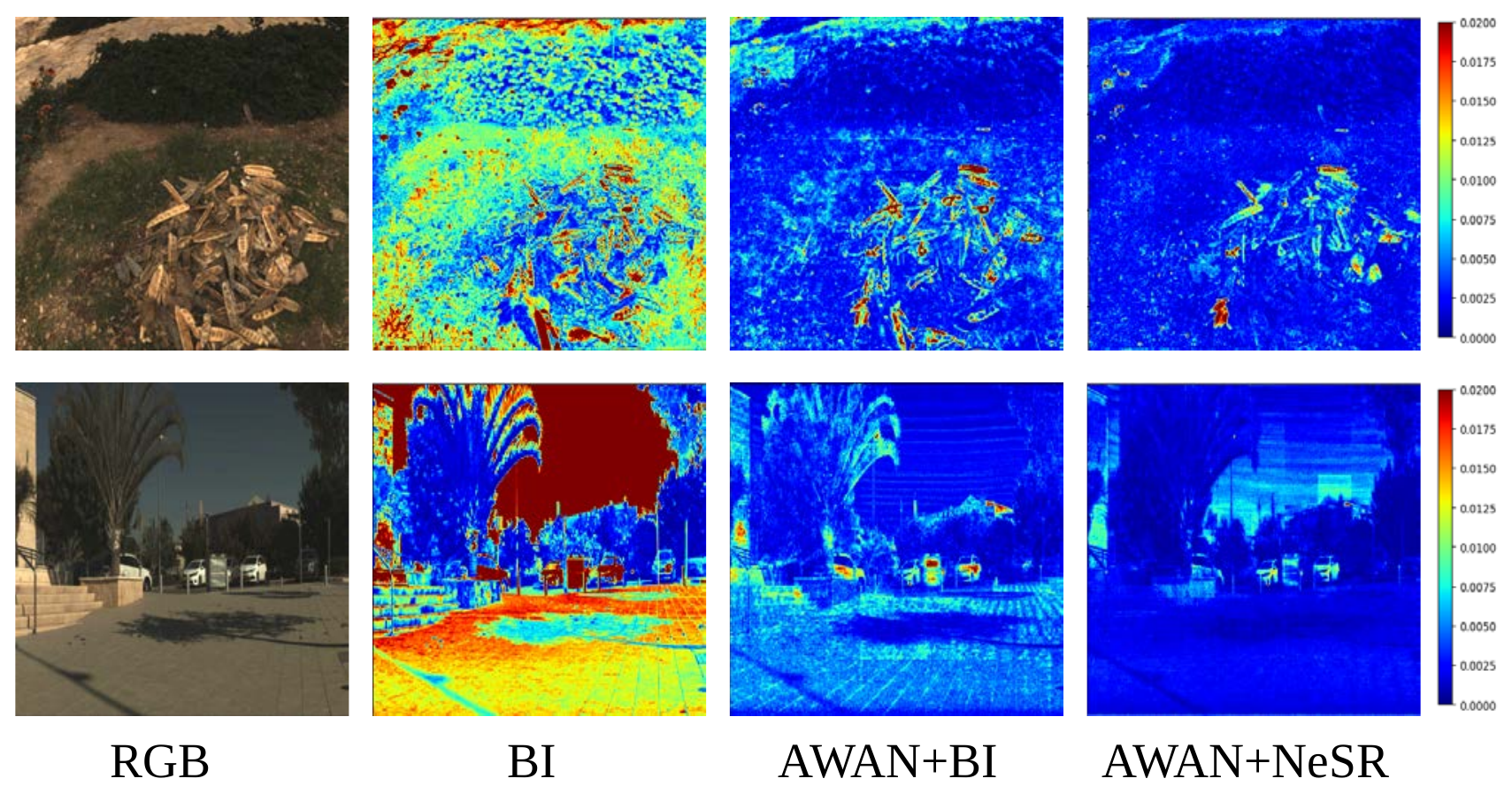}
  \caption{Error maps of reconstructed spectral images from RGB images on extreme spectral reconstruction.
  Top: Reconstructed results with 61 bands on ``BGU\_HS\_00177".
  Bottom: Reconstructed results with 51 bands on ``BGU\_HS\_00062".
  }
  \label{fig:outofdistribution}
\end{figure}

Our NeSR can also generate spectral images with an extreme band number beyond the training samples. We conduct experiments on the ICVL dataset to validate this point. During the training stage, NeSR takes an RGB image as the input, and the target spectral image is randomly sampled in the range of 7 to 31 bands.
In the inference stage, NeSR is able to reconstruct spectral images with an extreme band number (e.g., 61 bands), which have larger spectral resolution and are not contained in the training samples. 
We show the quantitative and qualitative comparison results in Table~\ref{tab:twoExp} and Figure~\ref{fig:outofdistribution}.

It can be seen that NeSR has superior performance on spectral image reconstruction with extreme band numbers and significantly outperforms the baselines.
It is worth noting that the reconstructed spectral images (41 bands to 61 bands) are out of the distribution of training samples, and NeSR has never seen them in the training stage, which indicates that NeSR empowers the practical continuous representation for spectral reconstruction.

\begin{table}[t]
\caption{Left: Quantitative results of reconstructing spectral images from RGB images on the extreme spectral reconstruction.
Right: Quantitative results of reconstructing high spectral-resolution images from low spectral-resolution images.
}
 \begin{minipage}[t]{0.49\textwidth}
  \centering
  
 \resizebox{1\linewidth}{!}{
       \renewcommand{\arraystretch}{1.4} 

\begin{tabular}{ccccccc}
\hline\hline
\multirow{2}{*}{Methods} & \multicolumn{2}{c}{61 bands} & \multicolumn{2}{c}{51 bands} & \multicolumn{2}{c}{41 bands} \\ \cline{2-7} 
                        & MRAE          & RMSE          & MRAE          & RMSE          & MRAE          & RMSE         \\ \hline
BI                      & 0.14844       & 0.02868       & 0.14962       & 0.02978       & 0.14914       & 0.03169      \\ \hline
AWAN + BI               & 0.03593       & 0.00848       & 0.03375       & 0.00791       & 0.03332       & 0.00771      \\ \hline
AWAN + NeSR             & 0.02689       & 0.00728       & 0.02655       & 0.00722       & 0.02677       & 0.00704      \\ \hline\hline
\end{tabular}
}
  \end{minipage}
  \hfill
 \begin{minipage}[t]{0.49\textwidth}
  \centering

   \resizebox{1\linewidth}{!}{
\begin{tabular}{ccccc}
\hline\hline
\multirow{2}{*}{Methods} & \multicolumn{2}{c}{16-31 bands} & \multicolumn{2}{c}{16-61 bands}  \\ \cline{2-5}
& MRAE & RMSE & MRAE & RMSE \\ \hline
BI  & 0.01521 & 0.00578 & 0.01837 & 0.00576\\ \hline
AWAN & 0.01285 & 0.00271 & 0.01346 & 0.00296 \\ \hline
AWAN + NeSR & 0.01232 & 0.00245 & 0.01312 & 0.00278    \\
\hline\hline
\end{tabular}
}
  \end{minipage}
 
\label{tab:twoExp}
\end{table}

\begin{figure}[!t]
\caption{Error maps of reconstructed spectral images on ICVL dataset.
  Top: Reconstructed results from 16 bands to 31 bands on ``BGU\_HS\_00061".
  Bottom: Reconstructed results from 16 bands to 61 bands on ``BGU\_HS\_00075".
  }
  \label{fig:spectral2spectral}
  \centering
  \includegraphics[width=0.85\linewidth]{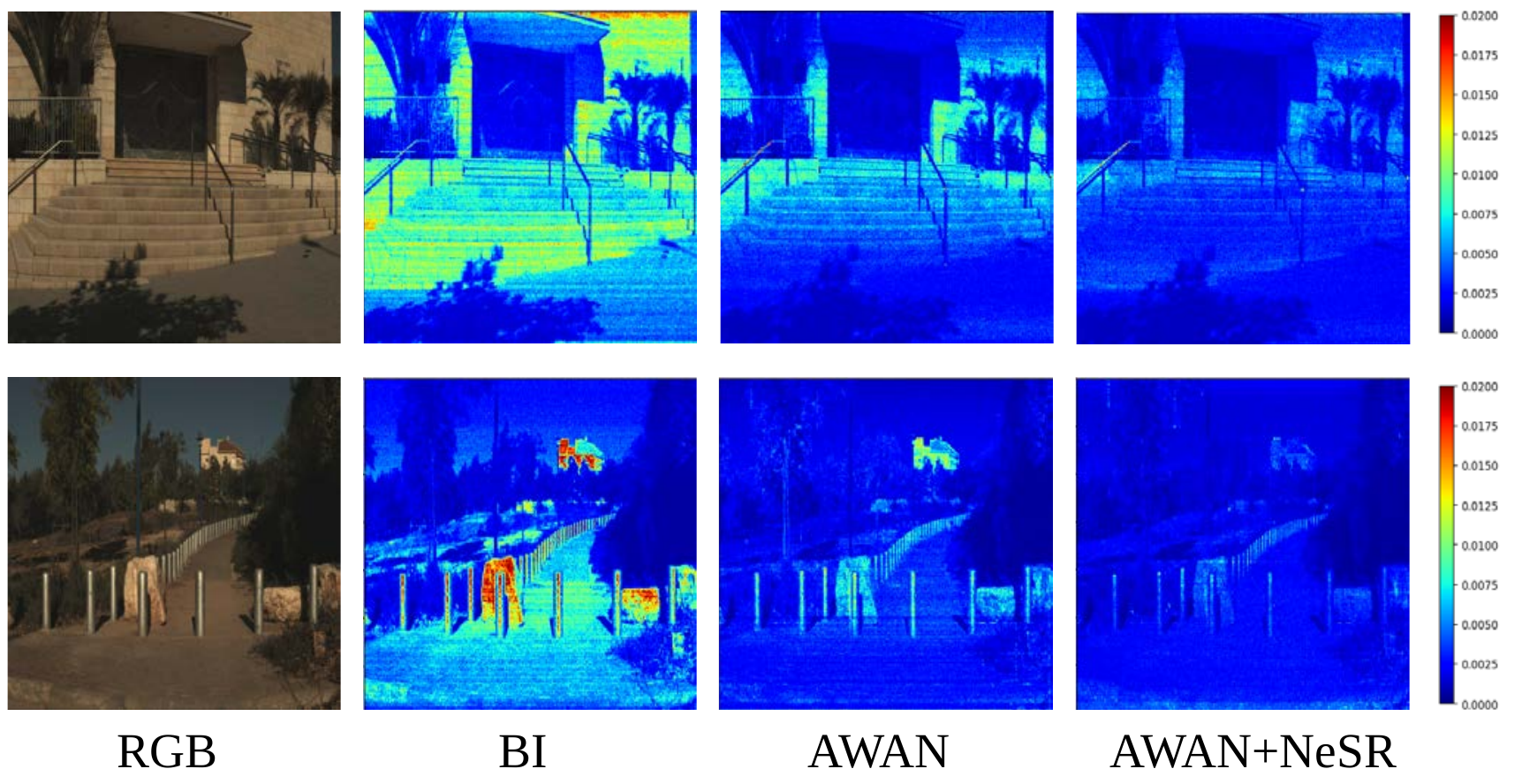}

\end{figure}

\subsection{Spectral Super Resolution}\label{sec:spectral2spectral}

Since NeSR aims to learn the continuous representation for reconstructing spectral images, it can be also applied to reconstruct high spectral-resolution images from low spectral-resolution images.
Specifically, we reconstruct the spectral image with 16 bands to 31 and 61 bands by a single model.
Experiments are conducted on the ICVL dataset, and BI and AWAN are selected as baseline methods, in which AWAN needs to train two times for different settings.
Other implementation details are the same as in Section~\ref{sota}.
As shown in Table~\ref{tab:twoExp}, our method (+NeSR) outperforms the baseline methods in both settings.
To give a visual comparison, we also show the error maps in both settings in Figure~\ref{fig:spectral2spectral}.
As can be seen, our method recovers the spectral information with a lower error.
The quantitative and qualitative comparison results verify that our method can be utilized for spectral super-resolution leveraging the continuous representation of spectral images.

\begin{table}[t]
\caption{Left: Ablation on SPI and NAM modules.
Right: Ablation on the attention mechanism. Both experiments are performed on the NTIRE2020 dataset.
}
\label{Tab:ablation}


 \begin{minipage}[t]{0.48\textwidth}
  \centering
  
 \resizebox{1\linewidth}{!}{
       \renewcommand{\arraystretch}{1.1} 

\begin{tabular}{ccc}
\hline\hline

Methods                   & MRAE  & RMSE   \\ \hline
AWAN   & {0.03441} & {0.01215} \\
AWAN+MLP                &  0.03387  & 0.01211         \\
AWAN+MLP+SPI        &  0.03139  & 0.01102     \\
AWAN+MLP+SPI+NAM         &  0.02996  &  0.00989     \\ 
\hline\hline

\end{tabular}
}
  \end{minipage}
 \begin{minipage}[t]{0.48\textwidth}
  \centering

   \resizebox{1\linewidth}{!}{
          \renewcommand{\arraystretch}{1.2} 

\begin{tabular}{ccc}
\hline\hline

Methods                   & MRAE  & RMSE   \\ \hline

Spectral-Wise                  &  0.03214  & 0.01001        \\

Spatial-Wise         &  0.03222  & 0.01032     \\

Spatial-Spectral-Wise         &  0.02996  & 0.00989     \\ 

\hline\hline

\end{tabular}
}
  \end{minipage}



\end{table}

\subsection{Ablation Studies}\label{subsec:ablation}


\begin{figure}[!t]

  \caption{Visualization results of ablation on the proposed modules on ``ARAD\_HS\_0464" of the NTIRE2020 dataset.}
  \label{fig:ablation}
  \centering
  \includegraphics[width=1\linewidth]{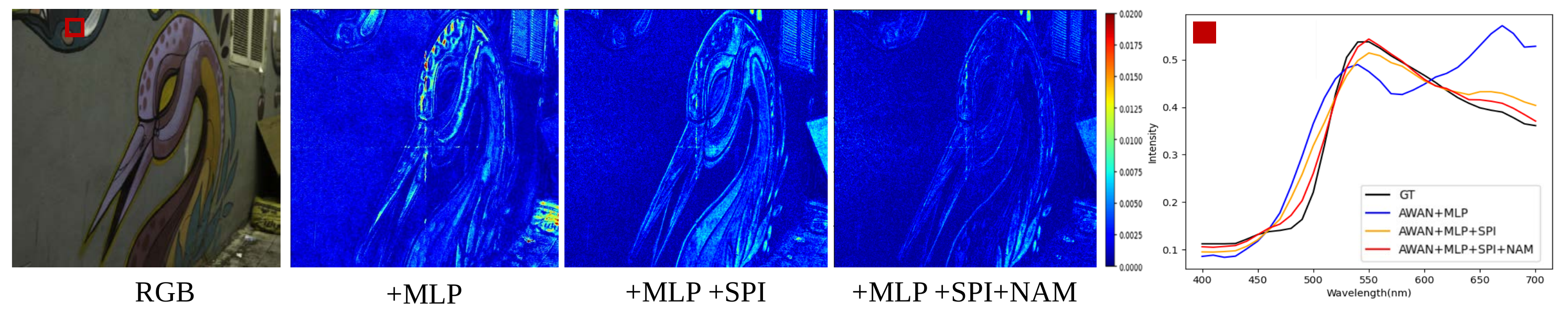}
%
\end{figure}

\textbf{Impact of SPI and NAM modules.}
We validate the effectiveness of the SPI module and the NAM module by adding them to the basic network step by step.
We use AWAN cascaded with an MLP as the basic network to ensure that it has a similar parameter amount to our method, and conduct an experiment on the NTIRE2020 ``Clean"  dataset.
The results of the ablation experiment are shown in Table~\ref{Tab:ablation}.
Since the deep feature lacks the spatial-spectral correlation, the baseline shows a limited reconstruction fidelity without the SPI and NAM modules.
When inserting the SPI module into the baseline, the MRAE decreases from 0.03387 to 0.03139, and the NAM module also contributes to 0.00123 MRAE decrease.
Moreover, from the error maps and the spectral curves in Figure~\ref{fig:ablation}, we can see that the reconstruction error decreases when inserting the SPI and NAM modules into the basic network.
The quantitative and qualitative comparison results demonstrate that our proposed modules are effective in enriching the spectral information and exploring the spatial-spectral correlation for learning the continuous spectral representation.

\textbf{Impact of the Attention Mechanism.}
To validate the effectiveness of our spectral-spatial-wise attention, we compare it with spectral-wise and spatial-wise attention~\cite{fu2019dual,zhang2018image}.
We use AWAN as the feature encoder and maintain the SPI module, and conduct an experiment on the NTIRE2020 ``Clean"  dataset.
For each model, we only replace the attention mechanism of the NAM module.
The results of the ablation experiment are shown in Table~\ref{Tab:ablation}.
Since the spatial-wise and spectral-wise attention cannot fully exploit the spectral-spatial correlation, they show a limited reconstruction accuracy.
Compared with the spatial-wise / spectral-wise attention, our spectral-spatial-wise attention decreases the MRAE from 0.03222 / 0.03214 to 0.02996.
The experiment results demonstrate that our nwe attention mechanism is effective in capturing the interactions of different channels for boosting performance.

%% file: 5_Conclusions.tex
\vspace{-0.3cm}
\section{Conclusion}
\vspace{-0.3cm}

In this paper, we propose NeSR to enable spectral reconstruction of arbitrary band numbers for the first time by learning the continuous spectral representation.
NeSR inputs an RGB image and a set of wavelengths, as the context of the scene and the target spectral coordinates, to control the output band numbers of the spectral image.
For high-fidelity reconstruction, we devise the SPI and NAM modules to exploit the spectral-spatial correlation in implicit neural representation. 
Extensive experiments demonstrate that NeSR significantly improves the reconstruction accuracy over the corresponding baselines with little parameter increase.
Moreover, NeSR can effectively reconstruct spectral images with an arbitrary and even extreme number of bands leveraging the continuous spectral representation.

\section*{Acknowledgments}
This work was supported in part by the National Natural Science Foundation of China under Grants 62131003 and 62021001.